\documentclass{article}

\usepackage{PRIMEarxiv}

\usepackage[utf8]{inputenc} % allow utf-8 input
\usepackage[T1]{fontenc}    % use 8-bit T1 fonts
\usepackage{hyperref}       % hyperlinks
\usepackage{url}            % simple URL typesetting
\usepackage{booktabs}       % professional-quality tables
\usepackage{amsfonts}       % blackboard math symbols
\usepackage{nicefrac}       % compact symbols for 1/2, etc.
\usepackage{microtype}      % microtypography
\usepackage{lipsum}
\usepackage{fancyhdr}       % header
\usepackage{graphicx}       % graphics
\graphicspath{{media/}}     % organize your images and other figures under media/ folder

\usepackage{multirow}
\newcommand*\rot{\rotatebox{90}}

\title{Mining United Nations General Assembly Debates\thanks{This work was funded by the European Union under the Horizon Europe grant OMINO (grant number 101086321) and by the Polish Ministry of Education and Science within the framework of the program titled International Projects Co-Financed.
(However, the views and opinions expressed are those of the authors only and do not necessarily reflect those of the European Union or the European Research Executive Agency. Neither the European Union nor the European Research Executive Agency can be held responsible for them.)
This research was also carried out with the support of the Faculty of Mathematics and Information Science at Warsaw University of Technology and its High-Performance Computing Center.}}

\author{
Mateusz Grzyb, Mateusz Krzyziński, 
  Bartłomiej Sobieski, Mikołaj Spytek, \\
  Warsaw University of Technology \\
  Faculty of Mathematics and Information Science \\
  Koszykowa 75, 00-662 Warsaw, Poland\\
  \texttt{\{mateusz.grzyb3, mateusz.krzyzinski, 
 bartlomiej.sobieski, mikolaj.spytek\}.stud@pw.edu.pl}
 \And 
 Bartosz Pieliński\\ 
 University of Warsaw \\
 Faculty of Political Science and International Studies \\
  Krakowskie Przedmieście 26/28, 00-927 Warsaw, Poland \\
\texttt{b.pielinski@uw.edu.pl}\\
 \And 
 Daniel Dan\\
 Modul University Vienna \\
  School of Applied Data Science\\
  Am Khalenberg 1, 1019, Vienna, Austria\\
\texttt{daniel.dan@modul.ac.at} \\
\And
 Anna Wróblewska\\
 Warsaw University of Technology \\ 
  Faculty of Mathematics and Information Science \\
  Koszykowa 75, 00-662 Warsaw, Poland\\
  \texttt{anna.wroblewska1@pw.edu.pl}\\
  }

\begin{document}
\maketitle

\begin{abstract}
This project explores the application of Natural Language Processing (NLP) techniques to analyse United Nations General Assembly (UNGA) speeches. Using NLP allows for the efficient processing and analysis of large volumes of textual data, enabling the extraction of semantic patterns, sentiment analysis, and topic modelling. 
Our goal is to deliver a comprehensive dataset and a tool (interface with descriptive statistics and automatically extracted topics) from which political scientists can derive insights into international relations and have the opportunity to have a nuanced understanding of global diplomatic discourse. 
\end{abstract}


\section{Introduction}
The United Nations (UN) is an international organization founded in 1945, comprising 193 member states. It was established after World War II with the intent to prevent future conflicts and foster global peace and security. The UN is a global forum where countries discuss and address critical issues ranging from international security, economic development, climate change, human rights, and humanitarian aid. It operates through various organs, including the General Assembly, the Security Council, and specialized agencies like UNESCO and WHO. The UN is pivotal in international cooperation and diplomacy, striving to maintain global stability and promote sustainable development.

The United Nations General Assembly (UNGA) serves as a global forum for member states to discuss and work together on international issues. UNGA provides a platform for multilateral discussion of the full spectrum of international issues covered by the UN Charter. 
It meets annually in regular sessions, with each member state having one vote. Essential functions include overseeing the UN's budget, appointing non-permanent members to the Security Council, and making recommendations in the form of General Assembly Resolutions.
The United Nations General Assembly (UNGA) debate transcripts are official records of the speeches delivered during the UNGA's annual sessions. These transcripts date from 1946 and provide detailed accounts of the statements made by representatives from each member state, reflecting their positions on various global issues. The speeches cover a wide range of topics, including international peace and security, economic development, human rights, environmental concerns, and other matters of global significance. 
The transcripts are essential for understanding the diplomatic stances of different countries, their interactions in the international arena, and the evolution of global policies and initiatives. %They are publicly accessible.

The huge volume of data source material makes manual analysis unfeasible. NLP methods came into play and allow extracting detailed information. We enriched the UNGA data with metadata and completed with the latest speeches. The main questions are: What topics are present in speeches? How does the language express the main international concerns? Are the date in time and the representative of a state a factor?
\begin{figure*}[htb]
    \centering
    \includegraphics[width=0.7\textwidth]{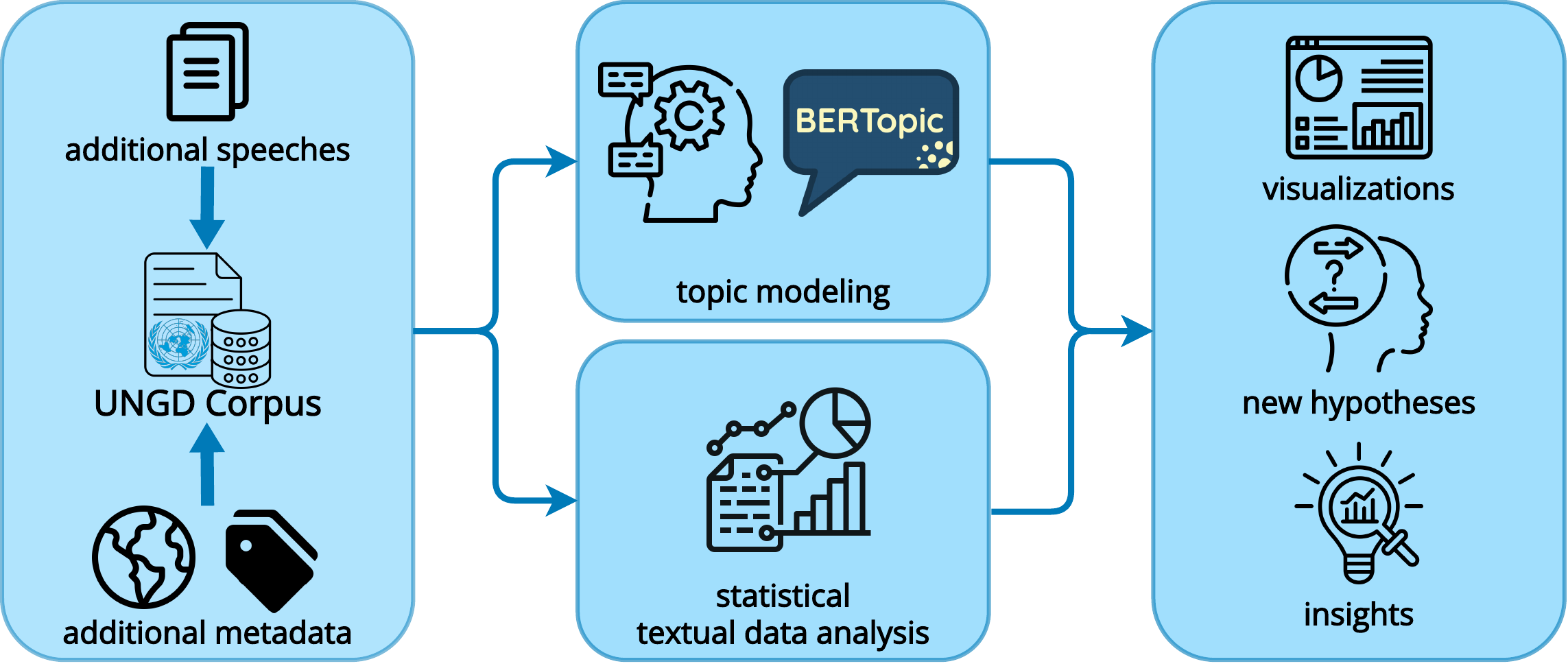}
    \caption{Diagram showcasing main steps of the project.}
    \label{fig:diagram}
\end{figure*}
Our project spanned over three main building blocks -- see Figure~\ref{fig:diagram}. The first one is data preparation, collecting the UNGA speeches between 1946-2023, completed with dates, names, the role of the speaker, and other additional features. The second is the exploration of the corpus and the calculation of speech statistics, together with preparing visualizations of the results. Third is the application of topic modelling techniques based on transformers.

\section{Related work}
The social and political science fields gained popularity lately due to the capability to process textual data \cite{hollibaugh2019use} and the NLP application was a game changer \cite{nay2018natural, glavavs2019computational}. The value of more than 10,000 speeches from representatives of 193 countries represents the most extensive resource of its kind. \cite{baturo2017understanding} provided additional examples of how such corpus can be leveraged; topic modeling is one of the techniques. In social science, countless examples of NLP methods are applied, especially topic modeling harnessed to compare structured corpora to find semantic similarities or dissimilarities \cite{valdez2021mining}. The most prominent technique in topic modelling is BERTopic \cite{bertopic}, which is considered state-of-the-art. As there are no current frameworks based on AI that allow for comparison analysis, this work would fill in this gap.

\section{Our approach}
%\subsection{Dataset preparation}
\noindent The dataset creation starts from an existing publicly available UNGD Corpus%, with more than 10,000 speeches stored in .txt files each
~\cite{ungd-corpus,Dasandi_Jankin_Baturo_2023}. The speeches are dated until 2022. The dataset was updated with the latest speeches extracted from the images through OCR software. %Metadata contained errors and inconsistencies, this needed to be corrected. The information was divided into regions and sub-regions, giving a multi-dimensional approach for further analysis. 
%\subsection{Dataset preparation}
%The preprocessing steps were done in Python and with the help of the spaCy package. After tokenization and lemmatization, the dataset was refined by excluding the stopwords. Other preprocessing steps, such as lowercasing were performed too.
%Initially, we updated this dataset by integrating newly delivered speeches from 2023, drawn from country-specific templates on the UN General Assembly website, as these speeches are not consolidated on the United Nations Digital Library.
Addressing data quality concerns, we correct errors and inconsistencies, particularly in the metadata. Moreover, our endeavor involves enriching the dataset's metadata by introducing additional variables and indices that provide a more comprehensive understanding of the characteristics of countries and their situation across different years. Furthermore, we integrate information from the United Nations geoscheme, providing division into regions and sub-regions. %This comprehensive approach ensures a robust and multi-dimensional characterization of countries, enriching the dataset for a more in-depth analysis of UNGA debates. 

%\subsection{Statistical analysis}
%Explorative statistics on the corpora was done to extract basic information. The most frequent words or bigrams, number of tokens, sentences and characters, although standard procedures, give basic insights. Readability metrics, part-of-speech proportions and document coherence values were also examined, yielding over 60 distinct statistics for each speech text.

%\subsection{BERTopic}
In order to deliver automatic tools for topic modelling, we introduced BERTopic \cite{bertopic}. %, which is a modular framework designed for topic modeling implementation using Python. 
It uses a pretrained Large Language Model, which uses semantic similarity between the words in documents. Because it uses word embeddings to generate topics, the semantic similarity of documents plays an important role in finding meaningful topics. 
In the main clustering process, BERTopic operates on uninterpretable word embeddings, and the human-understandable descriptions are extracted at the last step using the TF-IDF technique for each cluster. A metric is calculated for each word in each document, and the words with the highest scores are chosen as topic descriptions. 

%Drawbacks and limitations. The BERTopic model does not allow for a manual selection of the desired number of topics in the corpus. This can be fine-tuned by using HDBSCAN. Each document is assigned to one topic unlike in other approaches where each one is a mixture of topics. A high computational cost is a disadvantage, the usage of pretrained LLMs for getting embeddings for each document is time-consuming.

\section{Experiments and results}

\paragraph{Enhanced UNGD Corpus.}
The result of our work is the enhanced version of the United Nations General Debate Corpus, now inclusive of speeches delivered in 2023. %The incorporation of these speeches posed challenges due to their unavailability in the United Nations Digital Library, necessitating extraction from national templates, each unique to its respective country or, in some instances, absent. 
Consequently, we added 120 speeches, constituting a subset of the 195 originally delivered. In total, the enhanced UNGD Corpus consists of 10,679 speeches. 
The enhancement also improved metadata integrity, achieved through an iterative feature engineering process, including various refinements, such as rectifying typographical errors, harmonizing nomenclature to standardize varied representations, and addressing inaccuracies in ISO codes. Further data engineering and analyses were facilitated by resolving complications related to countries that no longer exist. %For instance, for countries with current equivalents, the current code was employed (e.g., Ukrainian SSR and Ukraine have \textit{UKR} code). In cases of countries that have split, their former codes were used (e.g., Yugoslavia -- YUG), with metadata alignment achieved through pertinent operations on the data pertaining to individual components of the erstwhile state, where applicable. Finally, \textcolor{blue}{the} letter casing is fixed and consistent for \textcolor{blue}{the} names of speakers and their positions. 
In total, 266 ISO codes, 1,862 country names, 3,877 names of speakers, and 3,199 of their positions were improved for enhanced accuracy.
A significant augmentation in the new version of the UNGD Corpus is the inclusion of additional metadata, comprising 10 new covariates matched to all speeches whenever applicable. %Table \ref{tab:metadata} details this data, along with its sources and coverage (percentage of documents for which they were matched). 
The data were drawn from diverse sources, including Gapminder, Our World in Data, and the United Nations, alongside its agencies and programs.\footnote{\url{https://www.gapminder.org/data/},\url{https://ourworldindata.org},\\\url{http://data.un.org/datamartinfo.aspx}}  %It is essential to acknowledge that certain covariates, such as HDI and Democracy Index, represent metrics that had not yet been introduced in 1946, contributing to inherent missing data.

\paragraph{BERTopic analyses.}
BERTopic was evaluated using different embedding methods. %, such as RoBERTa, and DistilBERT, sentence transformer models. 
The following metrics have been used for evaluating the topic models: (1) \textbf{Topic coherence}~\cite{coherence-1,coherence-2} metric utilizes various statistics drawn from the reference corpus and the split into topics, (2) \textbf{Topic diversity} \cite{diversity-1} which is a much simpler metric calculated based solely on the extracted topics and evaluating how much variability there is among them. 
Table~\ref{tab:coherence} and Table~\ref{tab:diversity} show values of the topic coherence and diversity metrics, respectively. %, all corresponding to a given topic modeling method and the number of topics combination. %It should be noted that values for the \texttt{roberta} method should be ignored because it always extracts only 6 distinct topics.

By both metrics, the LDA method performs significantly worse than methods from the BERTopic family. Regarding the topic coherence metric, the DistilBERT is the best for any number of topics. Regarding topic diversity metrics, the DistilBERT method is the first or second best, depending on the number of topics. %For this reason, we decide to utilize the \texttt{distilbert} method for our further analyses.
\begingroup
\renewcommand{\arraystretch}{1.2}
\begin{table*}[htb]
\small
\centering
\caption{Topic coherence metric results.}
\label{tab:coherence}
\begin{tabular}{|p{.2cm}|ccccc|}
\hline
\multirow{3}{*}{\rot{\textbf{\# topics}}} & \multicolumn{5}{c|}{\textbf{Topic modeling method}} \\ \cline{2-6} 
 & \multirow{2}{*}{LDA} & \multicolumn{4}{c|}{BERTopic} \\
 &  & all-MiniLM-L6-v2 & all-MiniLM-L12-v2 & all-mpnet-base-v2 & DistilBERT\\ \hline
10 & 0.387 & 0.458 & 0.432 & 0.433 & \textbf{0.463}  \\
20 & 0.397 & 0.450 & 0.422 & 0.427 & \textbf{0.499} \\
50 & 0.401 & 0.442 & 0.428 & 0.456 & \textbf{0.493}  \\ \hline
\end{tabular}
\end{table*}
\begin{table*}[htb]
\small
\centering
\caption{Topic diversity metric results.}
\label{tab:diversity}
\begin{tabular}{|p{.2cm}|ccccc|}
\hline
\multirow{3}{*}{\rot{\textbf{\# topics}}} & \multicolumn{5}{c|}{\textbf{Topic modeling method}} \\ \cline{2-6} 
 & \multirow{2}{*}{LDA} & \multicolumn{4}{c|}{BERTopic} \\
 &  & all-MiniLM-L6-v2 & all-MiniLM-L12-v2 & all-mpnet-base-v2 & DistilBERT \\ \hline
10 & 0.250 & \textbf{0.630} & 0.500 & 0.560 & 0.570  \\
20 & 0.250 & 0.445 & 0.430 & 0.450 & \textbf{0.470} \\
50 & 0.258 & 0.358 & 0.342 & \textbf{0.390} & 0.372 \\ \hline
\end{tabular}
\end{table*}

\endgroup

\paragraph{Visualisations.}
After choosing the methods, considering topic coherence and topic diversity, the results were summed up in an interactive application developed using the Streamlit data app framework. This app allows political science scholars to navigate the corpus efficiently and conduct analyses based on their domain of interest.\footnote{Our source code and links to binary dataset and our demo are available in  \href{https://github.com/grant-TraDA/NLP-2023W/tree/main/13. Mining UNGA debates/project1/solution}{our GitHub project}}

\section{Conclusions}
Our contributions can be summarized with a set of complementary approaches to enhancement and improvement of the UNGA debates corpus, development of an interactive application allowing easier extracting and summarising information, comprehensive tests on topic modelling using numerical metrics: BERTopic topic modeling pipeline applied together with 5 different embedding methods and compared with the LDA baseline using topic coherence and diversity metrics.

Notably, while the UN General Debates corpus was previously available, its utilization by non-technical users was hindered by limited ease of access. We believe that the refined user interface empowers researchers, policymakers, and other non-technical users, enabling them to explore and extract valuable insights from the data more efficiently and comprehensively than before.

%Bibliography
\bibliographystyle{unsrt}  
\bibliography{references}

\begin{thebibliography}{10}

\bibitem{hollibaugh2019use}
Gary~E Hollibaugh.
\newblock The use of text as data methods in public administration: A review and an application to agency priorities.
\newblock {\em Journal of Public Administration Research and Theory}, 2019.

\bibitem{nay2018natural}
John Nay.
\newblock {NLP} and machine learning for law and policy texts.
\newblock 2018.

\bibitem{glavavs2019computational}
Goran Glava{\v{s}}, Federico Nanni, and Simone~Paolo Ponzetto.
\newblock Computational analysis of political texts: bridging research efforts across communities.
\newblock In {\em ACL: Tutorials}, 2019.

\bibitem{baturo2017understanding}
Alexander Baturo, Niheer Dasandi, and Slava~J Mikhaylov.
\newblock Understanding state preferences with text as data: Introducing the un general debate corpus.
\newblock {\em Research \& Politics}, 2017.

\bibitem{valdez2021mining}
D.~Valdez, A.~C. Picket, B.~R. Young, and S.~Golden.
\newblock On mining words: the utility of topic models in health education research and practice.
\newblock {\em Health Promotion Practice}, 2021.

\bibitem{bertopic}
Maarten Grootendorst.
\newblock {BERTopic: Neural Topic Modeling with a Class-based TF-IDF Procedure}.
\newblock {\em arXiv:2203.05794}, 2022.

\bibitem{ungd-corpus}
Slava Jankin, Alexander Baturo, and Niheer Dasandi.
\newblock {\em {United Nations General Debate Corpus 1946-2022}}, 2017.

\bibitem{Dasandi_Jankin_Baturo_2023}
Niheer Dasandi, Slava Jankin, and Alexander Baturo.
\newblock Words to unite nations: The complete {UN General Debate Corpus}, 1946-present, May 2023.

\bibitem{coherence-1}
Jey~Han Lau, David Newman, and Timothy Baldwin.
\newblock Machine reading tea leaves: Automatically evaluating topic coherence and topic model quality.
\newblock In {\em EACL}, 2014.

\bibitem{coherence-2}
Michael R{\"o}der, Andreas Both, and Alexander Hinneburg.
\newblock Exploring the space of topic coherence measures.
\newblock In {\em WSDM}, 2015.

\bibitem{diversity-1}
Adji~B Dieng, Francisco~JR Ruiz, and David~M Blei.
\newblock Topic modeling in embedding spaces.
\newblock {\em TACL}, 2020.

\end{thebibliography}

\end{document}